\documentclass[journal]{vgtc}                     % final (journal style)
\onlineid{1866}

\vgtccategory{Research}

\vgtcpapertype{algorithm/technique}

\title{ Regularized Multi-Decoder Ensemble for an \\ Error-Aware Scene Representation Network }

\author{%
  Tianyu Xiong, \authororcid{Skylar W. Wurster}{0000-0001-6685-615X},
  Hanqi Guo, Tom Peterka, and Han-Wei Shen
}

\authorfooter{
  \item
  	Tianyu Xiong is with The Ohio State University. E-mail: xiong.336@osu.edu
}

\abstract{%
Feature grid Scene Representation Networks (SRNs) have been applied to scientific data as compact functional surrogates for analysis and visualization.
As SRNs are black-box lossy data representations, assessing the prediction quality is critical for scientific visualization applications to ensure that scientists can trust the information being visualized.  Currently, existing architectures do not support inference time reconstruction quality assessment, as coordinate-level errors cannot be evaluated in the absence of ground truth data.
By employing the uncertain neural network architecture in feature grid SRNs, we obtain prediction variances during inference time to facilitate confidence-aware data reconstruction. Specifically, we propose a parameter-efficient multi-decoder SRN (\textcolor{black}{MDSRN}) architecture consisting of a shared feature grid with multiple lightweight multi-layer perceptron decoders. \textcolor{black}{MDSRN} can generate a set of plausible predictions for a given input coordinate to compute the mean as the prediction \textcolor{black}{of the multi-decoder} ensemble and the variance as a confidence score.
The coordinate-level variance can be rendered along with the data to inform the reconstruction quality, or be integrated into uncertainty-aware volume visualization algorithms.
To prevent the misalignment between the quantified variance and the prediction quality, we propose a novel variance regularization loss for ensemble learning that promotes the Regularized multi-decoder SRN (R\textcolor{black}{MDSRN}) to obtain a more reliable variance that correlates closely to the true model error.
We comprehensively evaluate the quality of variance quantification and data reconstruction of Monte Carlo Dropout (MCD), Mean Field Variational Inference (MFVI), Deep Ensemble (DE), and Predicting Variance (PV) in comparison with our proposed \textcolor{black}{MDSRN} and R\textcolor{black}{MDSRN} applied to state-of-the-art feature grid SRNs across diverse scalar field datasets. We demonstrate that R\textcolor{black}{MDSRN} attains the most accurate data reconstruction and competitive variance-error correlation among uncertain SRNs under the same neural network parameter budgets. Furthermore, we present an adaptation of uncertainty-aware volume rendering and shed light on the potential of incorporating uncertain predictions in improving the quality of volume rendering for uncertain SRNs. Through ablation studies on the regularization strength and \textcolor{black}{decoder count}, we show that \textcolor{black}{MDSRN} and R\textcolor{black}{MDSRN} are expected to perform sufficiently well with a default configuration without requiring customized hyperparameter settings for different datasets.

}


\graphicspath{{figs/}{figures/}{pictures/}{images/}{./}} % where to search for the images

\usepackage[usenames,dvipsnames]{color}
\usepackage{tabu}                      % only used for the table example
\usepackage{booktabs}                  % only used for the table example
\usepackage{lipsum}                    % used to generate placeholder text
\usepackage{mwe}                       % used to generate placeholder figures
\usepackage{multirow}

\definecolor{gray1}{rgb}{0.78, 0.78, 0.78}
\definecolor{gray2}{rgb}{0.88, 0.88, 0.88}
\newcommand{\quotes}[1]{``#1''}

\usepackage{mathptmx}                  % use matching math font

\begin{document}

%%%%%%%%%%%%%%%%%%%%%%%%%%%%%%%%%%%%%%%%%%%%%%%%%%%%%%%%%%%%%%%%
%%%%%%%%%%%%%%%%%%%%%% START OF THE PAPER %%%%%%%%%%%%%%%%%%%%%%
%%%%%%%%%%%%%%%%%%%%%%%%%%%%%%%%%%%%%%%%%%%%%%%%%%%%%%%%%%%%%%%%


\maketitle

\section{Introduction}

Continuous functional representations of scientific datasets have gained attention as proxies for visualization and analysis thanks to their advantages in compactness, competitive modeling accuracy, and the ability to evaluate values and gradients at random locations without decoding the full volume. A Scene Representation Network (SRN) is a neural functional representation trained with volume data to learn a coordinate-value mapping to attain the aforementioned benefits. For the past several years, the scientific visualization (SciVis) community has improved SRN for volumetric data modeling in its compressiveness \cite{lu2021compressive, 10.2312:vmv.20221198, tang2023ecnr, gu2023nervi}, computational efficiency \cite{weiss2022fast, wu2023interactive, tang2023ecnr}, and adaptivity to data \cite{wurster2023adaptively}.

Despite the remarkable advancements in these areas, as a lossy approximation of the data, SRN lacks the ability to depict its prediction quality.
Since each value prediction made by SRN exhibits a certain level of error compared to the ground truth, and the most inaccurate predictions are often present in regions with scientific features that involve complex spatial patterns,
it is desirable to aid the process with a quantifiable metric that reflects how accurately the SRN reconstructs the data. 
However, it is non-trivial to obtain such a measurement of the coordinate-wise prediction quality at inference time 
because 
evaluating coordinate-wise prediction errors requires the ground truth data, which are often discarded after training. 
Alternatively, precomputing the error field and writing it to the disk would incur considerable storage costs equivalent to the data size and hence defeats the purpose of using an SRN as a compact surrogate.

In this paper, we propose a method to indicate the prediction quality for SRNs at inference time, which is storage-free and can be evaluated at arbitrary locations.
We explore the application of uncertain neural network architectures for computing prediction variances or uncertainties to indicate the confidence level of the model for its predictions.
These architectures, including Bayesian and ensemble methods\cite{gal2016dropout, blundell2015weight, lakshminarayanan2017simple}, can be used to make multiple plausible predictions for an input such that the variance and mean prediction amounts to a more accurate prediction with the error quantified. 
In addition to meeting the desired attributes for a proper prediction quality metric, the computed coordinate-level variance along with the mean can be conveniently granted a probabilistic interpretation and integrated into uncertainty-aware visualizations such as probabilistic marching cubes \cite{pothkow2011probabilistic} and uncertain volume rendering \cite{sakhaee2016statistical}, such that the uncertain predictions can be directly integrated to visualizations as an alternative to visualizing the mean predictions.

To empower SRNs with variance estimation for confidence-aware predictions, we introduce a feature grid SRN architecture with \textcolor{black}{an ensemble of decoders}, dubbed \textcolor{black}{multi-decoder SRN (\textcolor{black}{MDSRN})}, that can seamlessly extend existing feature grid SRNs to produce different plausible predictions for any given input coordinate. Common ensemble methods for variance quantification such as Deep Ensemble (DE) \cite{lakshminarayanan2017simple} require independent neural networks trained as members, thus resulting in a linear scaling of parameter counts to the number of members.
This might not be best suited for volumetric representations in SciVis where the compactness of the model is of great significance.
To better adapt ensembling to feature grid SRNs with improved parameter efficiency than DE, we propose a multi-decoder approach, and adapting our \textcolor{black}{MDSRN} to an existing feature grid SRN architecture simply requires training multiple MLP decoders along with a shared feature grid encoder.
Feature grid SRNs usually have the majority of parameters concentrated on the grid encoder, hence training additional lightweight MLPs only brings a negligible increase in the model size. Comparable approaches that can be applied to SRNs to generate uncertain predictions include the Bayesian neural network \cite{gal2016dropout, blundell2015weight} or directly predicting a variance. However, they can suffer from inferior data reconstruction accuracy, and we show the proposed \textcolor{black}{MDSRN} architecture achieves higher reconstruction accuracy with equivalent network size in \cref{sec:usrn_eval}.

Provided that the variance of predictions for each voxel serves as a metric for assessing the prediction quality, it is imperative for the variance to reflect the accuracy of the mean prediction, where 
voxels with high prediction errors should also possess high variances.
To ensure that the variance from \textcolor{black}{MDSRN} is a faithful representation of the prediction quality, we propose a variance regularization loss that synergizes with \textcolor{black}{MDSRN} to minimize the distribution discrepancy between the variance and error.
The network trained with the regularization, namely the Regularized \textcolor{black}{MDSRN} (R\textcolor{black}{MDSRN}), exhibits a superior quality of the acquired variance more similar to how error distributes across space.
With proper regularization strength, R\textcolor{black}{MDSRN} delivers the best data reconstruction and competitive variance quality as an uncertain SRN architecture for error-aware reconstruction and visualization.

To evaluate the efficacy of the proposed \textcolor{black}{MDSRN} architecture and the variance-regularized model R\textcolor{black}{MDSRN} in providing reliable prediction quality assessment for feature grid SRNs at inference time, we compare with Deep Ensemble (DE), Predicting Variance (PV), and popular Bayesian architectures including Monte Carlo Dropout (MCD) \cite{gal2016dropout} and Mean Field Variational Inference (MFVI) \cite{blundell2015weight}. We present both qualitative visualizations and quantitative results evaluating the spatial similarity of variance and prediction error of all architectures applied to state-of-the-art feature grid SRNs for SciVis \cite{weiss2022fast, wu2023interactive, wurster2023adaptively} across diverse datasets in addition to data reconstruction evaluations. Furthermore, we adapt an uncertainty-aware volume rendering algorithm \cite{sakhaee2016statistical} for uncertain SRNs and study the effect of incorporating uncertain predictions on the rendering quality.
To summarize, our contributions are threefold:

\begin{itemize}%[noitemsep,topsep=0pt]
    \item A parameter-efficient \textcolor{black}{architecture for feature grid SRN ensemble}, multi-decoder SRN (\textcolor{black}{MDSRN}), as a volumetric data representation, that enables post-training prediction quality assessment with coordinate-level prediction variance
    
    \item A novel variance regularization loss function used to train a Regularized \textcolor{black}{MDSRN} (R\textcolor{black}{MDSRN}) so that its prediction variance across space correlates more closely to the actual prediction error

    \item A comprehensive evaluation of uncertain neural network architectures applied to feature grid SRNs for SciVis. The evaluation spans both data reconstruction and variance estimation with qualitative and quantitative results. The six compared architectures include Monte Carlo Dropout (MCD), Mean Field Variational Inference (MFVI), Deep Ensemble (DE), Predicting Variance (PV), \textcolor{black}{MDSRN}, and R\textcolor{black}{MDSRN}.
    
\end{itemize}
% \vspace{-5 pt}

\begin{figure}[t]% specify a combination of t, b, p, or h for top, bottom, on its own page, or here
  \centering % avoid the use of \begin{center}...\end{center} and use \centering instead (more compact)
  \includegraphics[alt={overview and the pipeline of the RMDSRN}, width=\columnwidth]{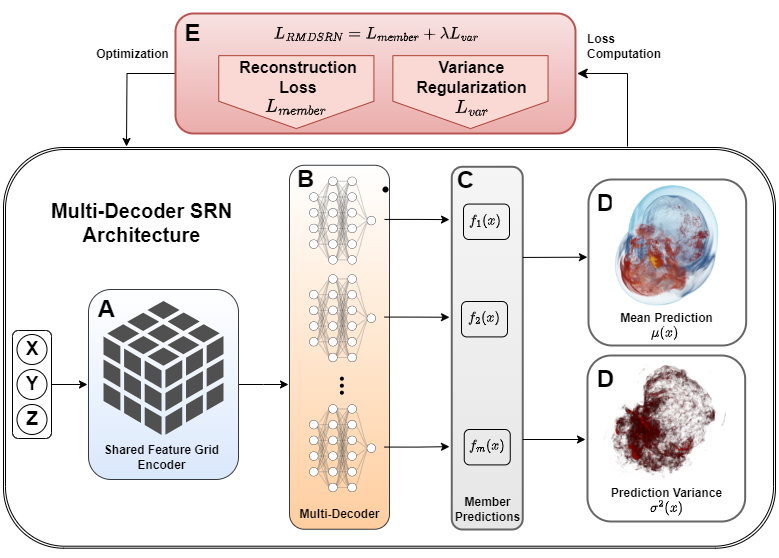}
  \caption{%
  	R\textcolor{black}{MDSRN} for volumetric data representation with prediction variance quantification for confidence assessment. \textcolor{black}{A feature grid SRN can be adapted to RMDSRN by training multiple decoders with the reconstruction loss introduced in \cref{sec:esrn} combing the weighted variance regularization loss detailed in \cref{sec:resrn} and \cref{sec:scheduler}.}
   }
  \label{fig:RESRN_overview}
  \vspace{-15 pt}
\end{figure}

\section{Related Work}

Our work targets adding error awareness to the predictions and visualizations of scene representation networks (SRNs) with uncertain neural network architectures, and we first review the fast-paced advancements of SRNs for scientific visualization, followed by related methods for uncertainty-aware neural networks.
% \vspace{-10 pt}

\subsection{Scene Representation Network for Scientific Visualization} \label{sec:related_srn}

Originally proposed as continuous representations of 3D shapes or scenes \cite{Park_2019_CVPR, sitzmann2019srns, mildenhall2020nerf},
scene representation networks (SRNs), or equivalently implicit neural representations (INRs), are applied to SciVis as continuous and compact surrogates that go beyond the discretized scientific data formats and support decoding the value at any arbitrary coordinate without full volume reconstruction.
A fully implicit approach of SRN that comprises multi-layer perceptrons (MLPs) for scientific data representation was first investigated by Lu et al. \cite{lu2021compressive}, and they proposed \textit{neurcomp} that extends the MLP with sinusoidal activation functions known as SIREN \cite{sitzmann2020implicit} with skip connections to achieve a high compression ratio for scalar field data in combination with parameter quantization.
SIREN-based SRNs were improved in various aspects following \textit{neurcomp}. Han and Wang \cite{han2022coordnet} showed the versatility of SRN in learning diverse tasks at both data and image levels for spatial-temporal datasets. To increase the computational efficiency of SIREN SRNs, Tang and Wang \cite{tang2023ecnr} introduced ECNR that extends the lightweight MLPs organized in a Laplacian pyramid proposed in MINER \cite{saragadam2022miner} to work on 4D scientific datasets with deep compression strategies for a high compression rate. Targeting the compression of volume visualization images for spatial-temporal datasets, Gu et al. \cite{gu2023nervi} proposed NeRVI that employs SIREN with skip connections and a CNN upscaling module for faster inference.

As the computational efficiency of MLP-based SRN can be concerning, a hybrid SRN architecture that connects an explicit feature grid with a lightweight MLP, namely feature grid SRN, attracted an interest in SciVis for fast SRNs.
Weiss et al. \cite{weiss2022fast} proposed fV-SRN with a composite encoder concatenating features from a dense feature grid and Fourier feature encoding \cite{mildenhall2020nerf, tancik2020fourfeat} followed by an MLP with SnakeAlt activation, and they further exploited GPU tensor cores for faster inference. Wu et al. \cite{wu2023interactive} achieved interactive training and visualization of hash grid SRN, known as NGP \cite{muller2022instant}, with an optimized training and rendering routine with out-of-core sampling and sample streaming for large-scale datasets. Höhlein et al. \cite{10.2312:vmv.20221198} explored the effectiveness of multi-grid and multi-decoder SRN in compressing meteorological ensembles. We note the motivation for our multi-decoder ensemble SRN differs from the network by Höhlein et al. as they use the different decoders to learn different volumes in an ensemble simulation dataset, whereas our method applies to a single scalar field for variance quantification as a realization of ensemble neural network. Farokhmanesh et al. \cite{10.2312:vmv.20231229} modeled bivariate correlations in ensemble datasets with a bipartite hash grid SRN dubbed NDF. For better generalization to unseen inputs, Wu et al. \cite{wu2023hyperinr} presented HyperINR with a hypernetwork on hash grids trained with knowledge distillation. As scientific data can exhibit certain degrees of sparsity, Wurster et al. \cite{wurster2023adaptively} proposed APMGSRN to concentrate SRN parameters to regions potentially containing scientific features with high errors through learned transformations for multiple feature grids.

\subsection{Uncertain Neural Network Architecture for Prediciton Variance Quantification} \label{sec:related_uq}

The critical necessity of understanding the trustworthiness of neural network predictions spurred a rich set of research on uncertainty-aware neural network architectures \cite{gawlikowski2023survey}.
\textcolor{black}{These methods can be applied to neural networks such that an uncertain prediction with variance quantification can be obtained for regression tasks.}
Ensemble and Bayesian Neural Network (BNN) are two prominent methods for this purpose. Monte Carlo Dropout (MCD) \cite{gal2016dropout} and Mean Field Variational Inference (MFVI) \cite{blundell2015weight} are two practical realizations of BNN. As for the ensemble approach, Deep Ensemble (DE) \cite{lakshminarayanan2017simple} advocated ensembling neural networks for uncertainty quantification with competitive quality compared to Bayesian methods. \textcolor{black}{Brief summaries of these uncertain neural network methods are included in \cref{sec:background} as they constitute competitive alternatives of variance quantification methods to be evaluated along with ours in \cref{sec:usrn_eval}.}
Since independent neural networks are trained, DE is parameter intensive, and there is a series of follow-up works focusing on parameter-efficient ensemble for neural networks \cite{valdenegro2019deep,Wen2020subens,turkoglu2022film,laurent:hal-04042523}.
Our work shares a similar idea to sub-ensemble \cite{valdenegro2019deep}, where efficiency is achieved by sharing parameters between members. 
While sub-ensemble is a generic method of parameter-sharing, we provide a specific strategy for the decomposition of the shared and unique parameters targeting feature grid SRNs.
In addition, our training routines also differ in ways that we train the shared feature grid jointly with all member decoders, whereas sub-ensemble uses sequential training of members with the shared component frozen after trained with the first member.

In the context of SRNs, several uncertainty-aware SRNs have been studied for different applications.
For Computed Tomography (CT) image representation, UncertaINR \cite{DBLP:journals/tmlr/VasconcelosHST23} proposed to use MCD for uncertain SRN. For neural radiance fields (NeRFs), a variance estimation technique used by ActiveNeRF \cite{pan2022activenerf} and U-NeRF \cite{chen2023leveraging} is to predict the variance as part of the network output in addition to a value prediction, and we refer to this as the Predicting Variance (PV) method as reviewed in \cref{sec:background_usrn}.
% treated as the mean of a Gaussian distribution, and the uncertain models optimize the negative log likelihood (NLL) of the predicted Gaussian. We refer to this method as Predicting Variance (PV) in \cref{sec:usrn_eval}.
S-NeRF \cite{9665942} proposed to learn independent logistic normal distributions of color and densities with variational inference, and CF-NeRF \cite{shen2022conditional} improved over S-NeRF with a conditional normalizing flow to learn a more flexible distribution of radiance field.
S\"{u}nderhauf et al. \cite{sunderhauf2023density} applied DE-based NeRF and formulated a density-aware uncertainty term to augment the RGB uncertainty.
Bayes’ Rays \cite{goli2023} proposed a Laplacian approximation of the parameter distribution of NeRF relying on a parametric perturbation field.
Despite different uncertain NeRFs being proposed, their training methods are often tailored to the NeRF pipeline, posing challenges to applications on other SRN tasks.

\begin{figure}[t]% specify a combination of t, b, p, or h for top, bottom, on its own page, or here
  \centering % avoid the use of \begin{center}...\end{center} and use \centering instead (more compact)
  \includegraphics[alt={illustration of the overconfident variance problem}, width=.75\columnwidth]{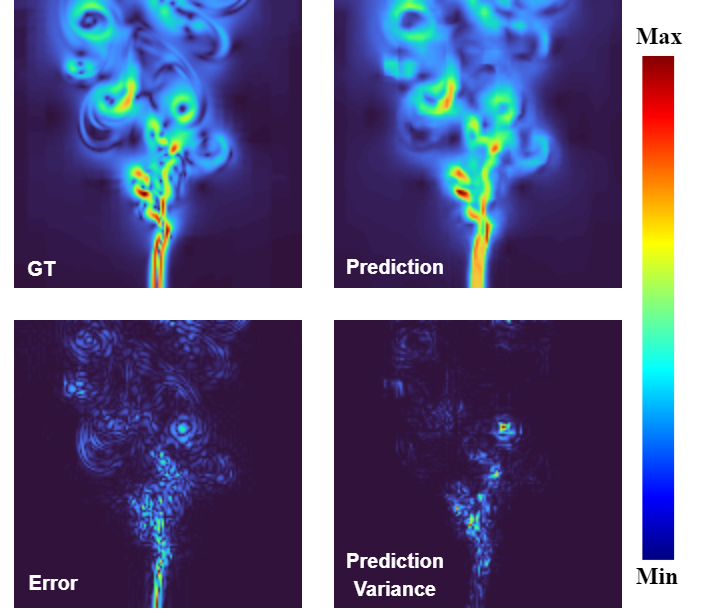}
  \caption{%
  	Illustration of the undesired overconfident variance problem of uncertain SRNs. Despite being the most inaccurate in the bottom thin structure as the error image shows, an uncertain SRN can fail to capture it with the highest variance elsewhere in the domain.
   }
  \label{fig:problem_uq}
  \vspace{-12 pt}
\end{figure}

{

\color{black}
{
\section{Background}
\label{sec:background}
As we propose to equip feature grid SRNs with prediction variance quantification to indicate the quality of the prediction, we briefly review SRNs with feature grid encoders to which our method in \cref{sec:method} is applied as well as the related variance quantification approaches for neural networks that are compared in \cref{sec:usrn_eval}.

\subsection{Feature Grid Scene Representation Networks}
\label{sec:background_srn}

Extending the general discussion of SRNs for SciVis in \cref{sec:related_srn}, we formally define SRNs and highlight important properties of its feature grid variants.
A scene representation network (SRN) is a neural network that predicts scalar or vector values from input coordinates,
and its architecture can be conceptually divided into two components, an encoder $E$ that transforms an input coordinate to a feature and a decoder $D$ that predicts the value given the feature. Formally, let $f$ be an SRN and $(x,y)$ be a coordinate-value mapping, the task of an SRN is to reconstruct $y$ given $x$: $f(x)=D(E(x))=y$.

An initial application of SRNs to scientific data was explored by Lu et al. \cite{lu2021compressive}. They proposed \textit{neurcomp} that uses multi-layer perceptrons (MLPs) for both the encoder and the decoder. Despite the high compressiveness achieved by \textit{neurcomp}, it is slow to evaluate as it consists of many wide fully connected layers. To improve the computational efficiency of SRNs, Weiss et al. \cite{weiss2022fast} proposed fV-SRN that substitutes the MLP encoder with a feature grid. A feature grid has learnable features defined on each voxel, and an input coordinate can be encoded with a trilinear-interpolated feature from neighboring voxels. The grid encoder can be evaluated more efficiently than an MLP which requires a chain of matrix multiplications. Furthermore, the decoder can also be made with fewer layers and neurons. As a result, the feature grid SRN can achieve considerable inference speedups. Following this conceptual model of grid encoders with lightweight decoders, more advanced feature grid SRNs are proposed to further improve the efficiency and accuracy for large-scale scientific data, such as Neural Graphics Primitives (NGP) \cite{muller2022instant, wu2023interactive} and the Adaptively Placed Multi-Grid SRN (APGMSRN) \cite{wurster2023adaptively}. In \cref{sec:method}, we introduce our approach that can add to feature grid SRNs the ability of reliable prediction variance quantification with a modification to the given network architecture along with a novel loss function, and our methods integrate well with state-of-the-art models like fV-SRN, NGP, and APMPSRN.

\subsection{Prediction Variance Quantification for Neural Networks}
\label{sec:background_usrn}

In addition to our work that proposes a customized strategy to modify the feature grid SRN architecture for variance quantification, there are alternative methods that can also be applied to achieve similar goals. Specifically, we briefly review Monte Carlo Dropout (MCD) \cite{gal2016dropout}, Mean Field Variational Inference (MFVI)\cite{blundell2015weight}, Deep Ensemble (DE)\cite{lakshminarayanan2017simple}, and Predicting Variance (PV)\cite{pan2022activenerf} before evaluating them in \cref{sec:usrn_eval}.

Similar to our methods, MCD, MFVI, and DE enable the network to produce multiple predictions to an input, from which a variance can be computed. Although dropout is initially proposed to reduce overfitting by randomly turning off neurons according to some probability \cite{srivastava2014dropout}, Gal and Ghahramani \cite{gal2016dropout} proposed to keep dropout at inference time such that different predictions can be acquired with multiple sets of network weights after dropout trails, and this MCD method bears a Bayesian neural network interpretation. MFVI \cite{blundell2015weight} is another Bayesian method, and it requires the network to learn a posterior Gaussian distribution for every weight, such that the network weight distributions can be sampled. Different samples of the network will in turn predict differently. DE \cite{lakshminarayanan2017simple} is a closely related method to ours as both approaches belong to the ensemble technique for uncertainty quantification. Instead of training one network, DE independently trains an ensemble of networks, such that the variance of the predictions from each member network can be calculated as uncertainty.

PV has a different realization of variance quantification compared to the abovementioned methods in that the network directly predicts the variance, and it was applied to SRNs, specifically neural radiance fields (NeRFs), by Pan et al.\cite{pan2022activenerf}. A PV network outputs both a mean and a variance prediction, effectively predicting a Gaussian distribution, and it is optimized with the negative log-likelihood (NLL) loss from the predicted Gaussian and the ground truth.
}

}

\section{Regularized Ensemble Scene Representation Network}
\label{sec:method}

Feature grid scene representation networks (SRNs) have received multitudes of improvements in computational efficiency and accuracy as surrogates for large-scale scientific data, yet there remains the question of where in the domain predictions can be trusted to be accurate such that the volume visualization will be in high quality with respect to the ground truth.

We explore extending feature grid SRNs with a network architecture that produces multiple predictions for every input to acquire prediction variance as a measurement of prediction quality and trustworthiness, which has been studied in diverse deep learning tasks \cite{eaton2018towards, scalia2020evaluating, ilg2018uncertainty, mukhoti2018evaluating, gustafsson2020evaluating}.
Our approach shown in \cref{fig:RESRN_overview} includes a parameter-efficient architecture for \textcolor{black}{an ensemble of} feature grid SRNs detailed in \cref{sec:esrn} as well as a novel variance regularization loss 
in \cref{sec:resrn}. \textcolor{black}{We then introduce the regularization strength scheduler that helps a more robust training to the choice of $\lambda$ in \cref{sec:scheduler}.}

\subsection{Multi-Decoder Ensemble SRN Architecture} \label{sec:esrn}

\textcolor{black}{Deep Ensemble (DE) \cite{lakshminarayanan2017simple} provides high-quality variance quantification and predictive performance to neural networks by training an ensemble of member networks to acquire diverse predictions.}
Despite the attractive performance of DE, an ensemble of independent member networks can be parameter-intensive \textcolor{black}{with a linear scaling factor to the number of members, which can lead to less optimal performance under a constrained parameter budget as in SRN tasks. To address the poor scaling of parameters of DE, we propose a parameter-efficient ensembling strategy for feature grid SRNs, dubbed multi-decoder SRN (MDSRN), that shares parameters between member networks.
When constructing an ensemble of feature grid SRNs,}
\textcolor{black}{all members of an MDSRN model share the feature grid}
so that none of them \textcolor{black}{learn} separate grids. Consequently, the one shared feature grid can use higher resolutions and larger feature sizes that all members can \textcolor{black}{benefit from}. Apart from the shared grid encoder, we do not share any layers in the multi-layer perceptron (MLP) decoder and instead use completely separate MLPs for different members. This is to prevent an excessive level of similarity between member networks, which can lead to a spatially homogenous variance that fails to distinguish regions that are challenging versus easy to learn.

We formally introduce the proposed \textcolor{black}{MDSRN} architecture 
\textcolor{black}{that adds variance quantification to feature grid SRNs. As illustrated in \cref{fig:RESRN_overview}, to incorporate our multi-decoder strategy to a feature grid SRN, while the encoder in part (B) requires no modification, an ensemble of decoders in part (C) needs to be trained to output multiple predictions such that a mean and a variance can be computed in part (D) for quality-informed reconstruction and visualization.}
\textcolor{black}{Following the notations in \cref{sec:background_srn}, for} an \textcolor{black}{MDSRN} containing $M$ member predictors 
that share the same encoder
with different decoders, denoted $f_i(x)=D_i(E(x))$ where $i=1, 2, .., M $, the mean prediction and the variance
\textcolor{black}{are computed as follows:}
% can be computed from the predictions of members:

\vspace{-7pt}
\begin{equation}
    \label{eq:ensemble_pred_var}
    \mu(x)=\frac{\sum^M_{i=1} f_i(x)}{M} \quad \quad \sigma^2(x)=\frac{\sum^M_{i=1}(f_i(x)-\mu(x))^2}{M-1}
\end{equation}

\textcolor{black}{The design is inspired by the observation that
% On the other hand, we empirically find
the bottleneck limiting the accuracy of an ensemble \textcolor{black}{of SRNs} is usually the capacity of the member networks. For example, for the same number of total parameters, an ensemble of 3 large independent SRNs often has a more accurate mean prediction than an ensemble of 5 smaller independent SRNs. This reveals a key to increasing the accuracy of an ensemble in a fixed compression level is to increase the capacity of member SRNs, and MDSRN achieves this through parameter-sharing.}
\textcolor{black}{The multi-decoder method}
can be applied easily to state-of-the-art feature grid SRNs for scientific data including APMGSRN \cite{wurster2023adaptively}, NGP\cite{muller2022instant, wu2023interactive}, and fV-SRN\cite{weiss2022fast} for post-training prediction confidence evaluation as \textcolor{black}{MDSRN} simply requires an ensemble of decoders of their specified architectures, such as MLP with ReLU activation for APMGSRN or SnakeAlt activation \cite{weiss2022fast} for fV-SRN, \textcolor{black}{despite that as a limitation of the ensemble method, training time can scale with the decoder members as shown in \cref{sec:usrn_eval}.}

\textcolor{black}{To optimize an \textcolor{black}{MDSRN} under the ensemble learning scheme similarly to DE networks,}
each member is supervised with the ground truth data to ensure they can make sufficiently plausible predictions. Let $x$ and $y$ be a coordinate-value pair and $B$ denote the number of pairs in a training batch, the per-member loss function 
\textcolor{black}{$L_{member}$ to train \textcolor{black}{MDSRN} in part (E) of \cref{fig:RESRN_overview}}
is defined as a sum of mean squared errors between each member's prediction and the ground truth:

\begin{equation}
    \label{eq:Lmembers}
    L_{\textcolor{black}{MDSRN}}= \textcolor{black}{L_{member}}  =\frac{1}{B} \sum^M_{i=1}\sum^B_{b=1}(f_i(x_b) - y_b)^2
\end{equation}

\subsection{Variance Regularization Loss Function} \label{sec:resrn}

Although \textcolor{black}{MDSRN} provides improved parameter efficiency for better data reconstruction accuracy than the conventional \textcolor{black}{network} ensemble approach at the same compression level, we found the spatial pattern of variance does not always align with that of error, as illustrated in \cref{fig:problem_uq}, and this can be problematic when the variance is utilized to evaluate the prediction quality. To mitigate the issue, our idea is to regularize the model with an additional loss function
% \textcolor{black}{$L_{var}$ as included in \cref{fig:RESRN_overview}}
so that the regularized \textcolor{black}{MDSRN} (R\textcolor{black}{MDSRN}) learns to promote higher similarity between the variance and error. In addition, the adjustments to variance should be ideally based on its own scale to avoid equating two different quantities with potentially disparate value ranges.

We propose to minimize the dissimilarity of the variance and error in the context of probability distributions to achieve scale-invariant matching. By normalizing the variance and error fields such that both quantities integrate to one over the spatial domain, we can obtain valid probability density functions (PDFs) in 3D. Furthermore, since the density of a value at a location indicates the relative magnitude of the value compared with others over the domain, locations with overconfident variance can be naturally identified by those with lower variance densities than the error densities. Consequently, minimizing the difference between the two density functions effectively adjusts the variance in a scale-invariant way.

To acquire proper densities for the variance and error, we can scale each of the values by their total aggregates in the volume. Formally, let $X$ denote all coordinates encompassed by the volume extent and $v(x)=y$ be a function that returns the ground truth value at a location, the variance density function $f_{\sigma^2}$ and the error density function $f_{\delta}$ are defined as:
\vspace{-12pt}

\begin{equation}
    \label{eq:f_var}
    f_{\sigma^2}(x)= \frac{\sigma^2(x)}{ \int_X \sigma^2(x_j)dx_j} \quad \quad
    f_{\delta}(x)= \frac{( \mu(x)- v(x) )^2}{\int_X ( \mu(x_j)-v(x_j) )^2dx_j}
\end{equation}

Although \cref{eq:f_var} describes the density functions in the continuous space over the volume for SRNs as continuous representations, only random voxel samples are available in a training batch from an optimization step.
% One option is to delay backpropagations until all voxels are fed to the network to obtain the variance and error densities in the whole volume, but this is inefficient for training.
To compute the densities during training, we normalize the variances and errors of the sampled points in a batch by their respective sums, which is equivalent to replacing the integrals in \cref{eq:f_var} by a summation over voxels in a batch, such that the densities constitute discrete approximations of \cref{eq:f_var}. The batch-wise densities are fast to compute and can update the model parameters for every optimization step to work with the member reconstruction loss, and results in \cref{tab:acc_uq} show this approach can effectively align the variance of R\textcolor{black}{MDSRN} with the error across the volume without increasing the training time significantly compared to \textcolor{black}{MDSRN}.

After density functions are defined, we minimize their difference with Kullback–Leibler (KL) divergence as a measure of dissimilarity between probability distributions, and \cref{eq:Lvar} shows the variance regularization $L_{var}$ used during training. KL divergence especially helps R\textcolor{black}{MDSRN} prioritize matching the variance for high-error regions that can greatly affect the accuracy of visualizations, and it is accomplished by weighing the log-space density difference by the error density, hence points with high errors contribute more to the loss. We note that gradients incurred in the computation of error densities are not recorded, because we would like member models to minimize KL divergence solely by steering the variance of their predictions toward similar patterns to error, so gradients should only flow from the loss function to the variance densities.
% \vspace{-5pt}

\begin{equation}
    \label{eq:Lvar}
    L_{var}=\frac{1}{B} \sum^B_{j=1}f_\delta(x_j)log(\frac{f_\delta(x_j)}{f_{\sigma^2}(x_j)})
\end{equation}

\textcolor{black}{
By adapting our multi-decoder design and the variance-regularized training as \cref{fig:RESRN_overview} presents, feature grid SRN can achieve reliable variance quantification.
The loss function of the final Regularized \textcolor{black}{MDSRN} (R\textcolor{black}{MDSRN})
becomes a weighted sum of the member loss and the variance regularization
}
: $L_{R\textcolor{black}{MDSRN}}=\textcolor{black}{L_{member}}+\lambda L_{var}$.

\begin{figure}[t]% specify a combination of t, b, p, or h for top, bottom, on its own page, or here
  \centering
  \includegraphics[alt={a line chart visualizing the variance regularization strength throughout training}, width=.75\columnwidth]{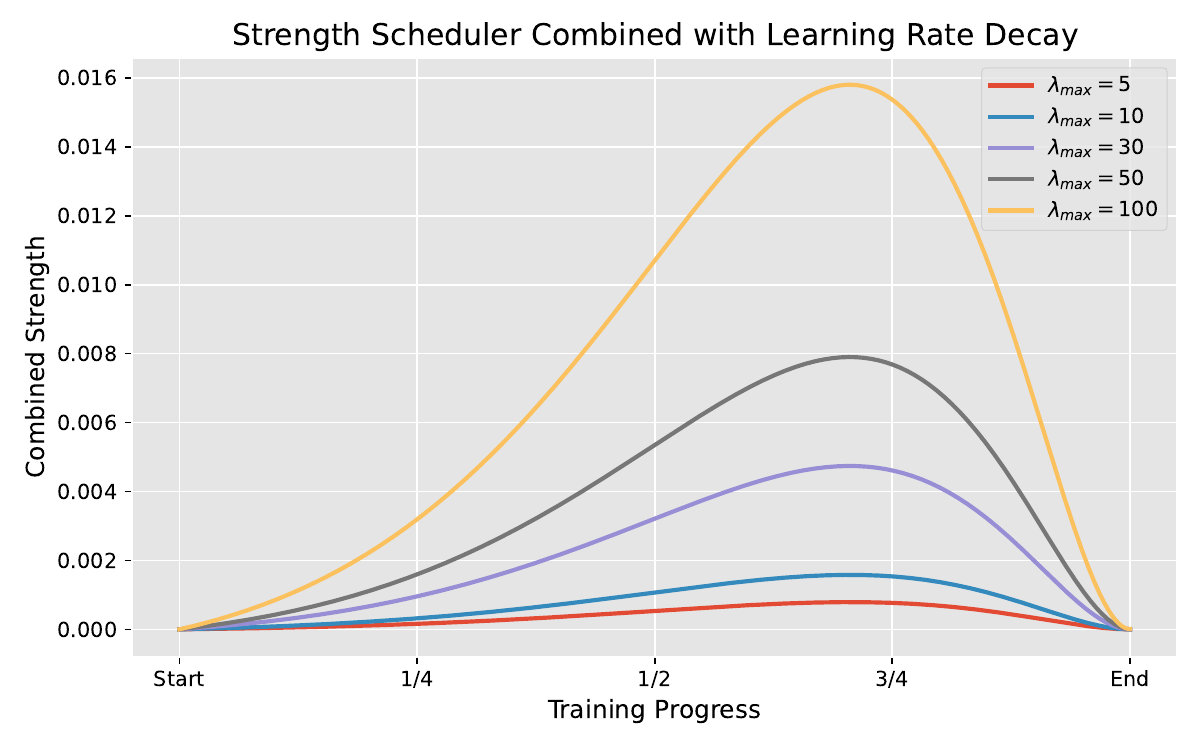}
  \caption{%
  	Visualization of the combined variance regularization strength, which is the learning rate decayed with cosine annealing multiplied by $\lambda(t)$ from \cref{eq:Lvar_scheduler}, at each step of training with varying $\lambda_{max}$ for the scheduler.
   }
  \label{fig:scheduler}
  \vspace{-15pt}
\end{figure}

\begin{figure*}[t]
\centering
  \includegraphics[alt={volume renderings of reconstructed volumes}, width=\textwidth]{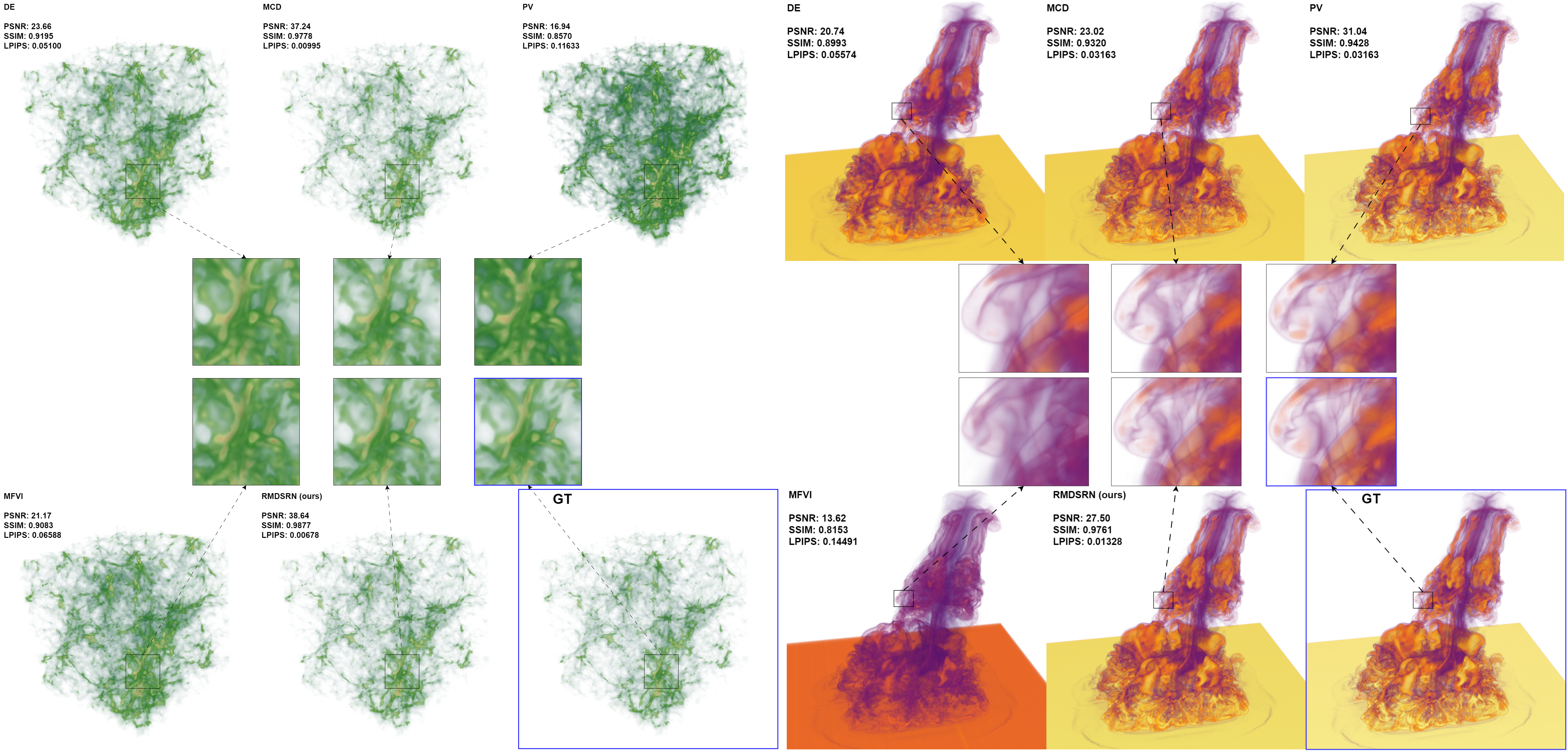}
  \caption{
    Volume renderings of reconstructed data for Nyx (left) and Asteroid (Right) as well as the ground truth in blue borders.  Both image-level evaluation metrics and the enlarged views show the renderings from R\textcolor{black}{MDSRN} maintain the highest fidelity overall and reproduce feature structures the most clearly as evidenced by SSIM and the perceptual loss.
  }
  \label{fig:Reconstruction}
\vspace{-18pt}  
\end{figure*}

\subsection{Training Regularized \textcolor{black}{MDSRN} with \textcolor{black}{Exponential} Growth Weight Scheduler} \label{sec:scheduler}

\textcolor{black}{Compound loss functions usually benefit from appropriate weights on each term for the best results, such as in image generation networks like InSituNet \cite{he2019insitunet} and FoVolNet \cite{bauer2022fovolnet}.
The quality of a trained RMDSRN model is also dependent on a proper regularization strength $\lambda$. We use a $\lambda$ scheduler to provide more robust training without requiring a hyperparameter search as the detailed ablation study in \cref{sec:abla_lambda} demonstrates. The R\textcolor{black}{MDSRN} trained with the scheduler exhibits improved overall performance than the unregularized MDSRN for a wide $\lambda$ range as well as the default recommendation.
}

\textcolor{black}{The scheduler defined in \cref{eq:Lvar_scheduler} gradually increases $\lambda$ from a minimal value to a maximum at an exponential rate during training.}
The principle \textcolor{black}{behind} our $\lambda$ scheduling is to silence the regularization in early iterations and scale it up as the model approaches convergence in the member loss.
The motivation is that since model predictions and hence errors will change after each optimization step, distribution matching becomes a moving target.
On the other hand, the true targets that the variance should learn to correlate are the errors in later training iterations, as they do not vary significantly and are more consistent representations of the final R\textcolor{black}{MDSRN} error at inference time.
With this principle, our scheduler starts $\lambda$ with a small value ideally close to zero and keeps it small in the early iterations of training to allow a faster progression of the model accuracy. Following exponential growth rates, in later stages of training, the scheduler can quickly raise $\lambda$ to allow a satisfactory fit to the final error pattern.
The weight on training step $t$ is defined as:
\vspace{-5pt}

\begin{equation}
    \label{eq:Lvar_scheduler}
    \lambda(t) = \lambda_{min} + (\lambda_{max}-\lambda_{min})\frac{r^{\frac{t-1}{t_{max-1}}}-1}{r-1}
\end{equation}

At step 1, the weight is the minimum value $\lambda_{min}$ and it increases at an exponential rate to $\lambda_{max}$ at the last training step $t_{max}$. Here $\lambda_{max}$ is a hyperparameter to be tuned for the best result of variance regularization, while for $\lambda_{min}$ a default of zero works generally. The parameter $r\in(1,\infty]$ controls the growth rate of the scheduler.
Rates close to 1 result in linear-like growths and the higher the rate, the steeper the growth will be in later iterations and the less weight for early iterations.

As the strength scheduler is applied jointly with learning rate decay, the actual weight on the regularization is $\lambda(t)$ in \cref{eq:Lvar_scheduler} multiplied by the current learning rate. We visualize in \cref{fig:scheduler} the combined strength throughout training with a cosine annealing learning rate scheduler \cite{DBLP:conf/iclr/LoshchilovH17} used in \cref{sec:usrn_eval} with an initial learning rate of 5.0e-3. For all $\lambda_{max}$ shown, $\lambda_{min}$ and $r$ are set to 0 and 500. The combined strength grows slowly in early training and progresses quickly to a maximum before 3/4 of training, followed by decreasing values until the end of training. This follows our motivation to scale $\lambda$ up in later training stages, and the shrinking $\lambda$ towards the end helps the regularization converge.

In summary, $\lambda_{max}$ of the scheduler \textcolor{black}{can} be tuned for the best results from variance regularization. For $\lambda_{min}$, setting it to zero should fit most training scenarios. Regarding the growth rate $r$, we recommend a default $r$ of 500 so the weight does not grow significantly until around 1/4 of the total training steps as visualized in \cref{fig:scheduler}. \textcolor{black}{ A detailed ablation study of $\lambda_{max}$ and the scheduler can be found in \cref{sec:abla_lambda}.}

\begin{table}[t]
\centering
\renewcommand{\arraystretch}{1.2}
\caption{
Quantitative data reconstruction and variance evaluation results. For each dataset, evaluated methods include the non-uncertain SRN (as in \quotes{SRN}) and uncertain models with a specified base SRN architecture. All models in the same dataset have the same size as indicated in the Base SRN column. PSNR evaluates data-level reconstruction accuracy. Pearson correlation (corr) and our modified Jaccard Index with spatial tolerance (JI-ST) assess the spatial similarity between the prediction variance and error. Negative log likelihood (NLL) evaluates the quality of Gaussian distributions parameterized by prediction means and variances in representing the data. The top two statistics among uncertain models in each dataset are bolded with the best one underlined.
}
\Huge

\label{tab:acc_uq}
\resizebox{\columnwidth}{!}{%
\begin{tabular}{c|c|cccccc}
\multicolumn{1}{l|}{Dataset} & Base SRN & Model & \multicolumn{1}{c}{PSNR (dB) $\uparrow$} & \multicolumn{1}{c}{Corr $\uparrow$} & \multicolumn{1}{c}{JS-ST(1\%/5\%) $\uparrow$} & \multicolumn{1}{c}{NLL $\downarrow$} & \multicolumn{1}{c}{Training time} \\ \hline 
\multirow{6}{*}{\shortstack{ \\ \\\\ \\ \\ \\ \\ \\ \\ \\ \\ \\ \\ Plume \\ \\ \\ $512 \times 128^2$ \\ \\ \\ 32 MB }} & \multirow{5}{*}{\shortstack{ \\ \\ \\ \\ \\ \\ \\ \\ \\ \\ \\ \\ APMGSRN \\ \\ \\  \\ 0.35 MB}}
 & SRN & 49.66 & N/A & N/A & N/A & 404s \\
 &  & MCD & 44.73 & 0.2928 & 43.9\% / 54.9\% & 8.28 & \textbf{430s} \\
 &  & MFVI & 46.00 & 0.3172 & 47.0\% / 50.1\% & -0.59 & 495s \\
 &  & DE & 46.07 & 0.3726 & 40.8\% / 52.9\% & 5.86 & 804s \\
 &  & PV & 42.53 & \textbf{0.5415} & \underline{\textbf{61.9\% / 76.9\%}} & \underline{\textbf{-5.88}} & \underline{\textbf{423s}} \\
 &  & \textcolor{black}{MDSRN} & \textbf{46.66} & 0.3521 & 44.6\% / 59.5\% & 759.08 & 760s \\
 &  & R\textcolor{black}{MDSRN} & \underline{\textbf{47.58}} & \underline{\textbf{0.6154}} & \textbf{60.1\% / 64.3\%} & \textbf{-4.27} & 780s \\ \hline

\multirow{6}{*}{\shortstack{ \\ \\\\ \\ \\ \\ \\ \\ \\ \\ \\ \\ \\ Nyx \\ $256^3$ \\\\\\ 64 MB }} & \multirow{5}{*}{\shortstack{ \\ \\ \\ \\ \\ \\ \\ \\ \\ \\ \\ \\ fV-SRN \\ \\ \\ \\ 3.21 MB}}
 & SRN & 39.39 & N/A & N/A & N/A & 346s \\
 &  & MCD & 37.09 & 0.1353 & 25.7\% / 34.1\% & 6.42 & \underline{\textbf{386s}} \\
 &  & MFVI & 37.70 & 0.0963 & 20.6\% / 30.8\% & 3.46 & 494s \\
 &  & DE & 33.98 & 0.0702 & 12.5\% / 24.9\% & -0.85 & 789s \\
 &  & PV & 37.33 & \underline{\textbf{0.4536}} & \underline{\textbf{50.5\% / 54.3\%}} & \underline{\textbf{-3.06}} & \textbf{402s} \\
 &  & \textcolor{black}{MDSRN} & \textbf{39.01} & 0.0215 & 11.3\% / 27.9\% & 418.96 & 730s \\
 &  & R\textcolor{black}{MDSRN} & \underline{\textbf{39.04}} & \textbf{0.2134} & \textbf{27.7\% / 41.4\%} & \textbf{-1.16} & 746s \\ \hline

\multirow{6}{*}{\shortstack{ \\ \\\\ \\ \\ \\ \\ \\ \\ \\ \\ \\ \\ Supernova \\ $432^3$ \\\\\\ 308 MB }} & \multirow{5}{*}{\shortstack{ \\ \\ \\ \\ \\ \\ \\ \\ \\ \\ \\ \\ APMGSRN \\ \\ \\ \\ 4.33 MB}}
 & SRN & 50.36 & N/A & N/A & N/A & 346s \\
 &  & MCD & 43.52 & 0.3578 & 38.9\% / 59.4\% & -3.72 & \underline{\textbf{386s}} \\
 &  & MFVI & 46.83 & 0.2505 & 32.8\% / 55.9\% & -3.98 & 313s \\
 &  & DE & 46.76 & 0.4268 & 47.6\% / 60.3\% & \underline{\textbf{-7.48}} & 789s \\
 &  & PV & 46.01 & \underline{\textbf{0.5364}} & \underline{\textbf{64.6\% / 74.0\%}} & -6.46 & \textbf{390s} \\
 &  & \textcolor{black}{MDSRN} & \textbf{49.31} & 0.2089 & 32.2\% / 49.2\% & 25.66 & 730s \\
 &  & R\textcolor{black}{MDSRN} & \underline{\textbf{49.35}} & \textbf{0.5026} & \textbf{59.3\% / 68.8\%} & \textbf{-7.05} & 746s \\ \hline

\multirow{6}{*}{\shortstack{ \\ \\\\ \\ \\ \\ \\ \\ \\ \\ \\ \\ \\ Asteroid \\\\\\ $1000^3$ \\\\\\ 3815 MB }} & \multirow{5}{*}{\shortstack{ \\ \\ \\ \\ \\ \\ \\ \\ \\ \\ \\ \\ \\ NGP \\ \\ \\ \\ 40.21 MB}}
 & SRN & 45.71 & N/A & N/A & N/A & 188s \\
 &  & MCD & 39.64 & 0.1242 & 19.6\% / 56.7\% & -3.40 & \textbf{209s} \\
 &  & MFVI & 40.85 & 0.1609 & 31.6\% / 44.4\% & -4.10 & 417s \\
 &  & DE & 36.19 & 0.4575 & 66.2\% / 91.2\% & \underline{\textbf{-7.87}} & 405s \\
 &  & PV & 38.43 & \underline{\textbf{0.6213}} & \underline{\textbf{79.4\%}} / \textbf{87.3\%} & -7.51 & \underline{\textbf{192s}} \\
 &  & \textcolor{black}{MDSRN} & \textbf{45.86} & 0.2455 & 65.4 \% / 90.8\% & -4.57 & 372s \\
 &  & R\textcolor{black}{MDSRN} & \underline{\textbf{46.12}} & \textbf{0.4618} & \textbf{73.5\% / \underline{91.6\%}} & \textbf{-7.59} & 345s \\ \hline

\multirow{6}{*}{\shortstack{ \\ \\\\ \\ \\ \\ \\ \\ \\ \\ \\ \\ \\ Isotropic \\ $1024^3$ \\\\\\ 4096 MB }} & \multirow{5}{*}{\shortstack{ \\ \\ \\ \\ \\ \\ \\ \\ \\ \\ \\ \\ fV-SRN \\ \\ \\ \\ 40.21 MB}}
 & SRN & 39.57 & N/A & N/A & N/A & 407s \\
 &  & MCD & 36.44 & 0.0738 & 10.2\% / 20.9\% & 7.82 & \textbf{440s} \\
 &  & MFVI & 34.31 & 0.0373 &  5.4\% / 12.6\% & 4.89 & 529s \\
 &  & DE & 34.99 & 0.2083 & 16.4\% / 26.9\% & -0.33 & 813s \\
 &  & PV & 36.99 & \underline{\textbf{0.4901}} & \underline{\textbf{44.2\% / 55.0\%}} & \underline{\textbf{-3.24}} & \underline{\textbf{411s}} \\
 &  & \textcolor{black}{MDSRN} & \underline{\textbf{39.48}} & 0.0200 &  7.6\% / 18.1\% & 1616.72 & 776s \\
 &  & R\textcolor{black}{MDSRN} & \underline{\textbf{39.48}} & \textbf{0.2876} & \textbf{29.8\% / 41.4\%}  & \textbf{-1.68} & 790s \\

\end{tabular}
}
\vspace{-5mm}
\end{table}

\section{Experiments} \label{sec:experiments}
To study the performance of uncertain neural network architectures on feature grid SRNs for scientific data, we evaluate \textcolor{black}{MDSRN}, R\textcolor{black}{MDSRN}, Deep Ensemble (DE) \cite{lakshminarayanan2017simple}, Monte Carlo Dropout (MCD) \cite{gal2016dropout}, Mean Field Variational Inference (MFVI) \cite{blundell2015weight}, and Predicting Variance (PV) in reconstruction accuracy and variance quality with evaluation metrics designed for volume visualization along with qualitative results in \cref{sec:usrn_eval}. \textcolor{black}{Related methods are reviewed in \cref{sec:background_usrn}.} We then study an adaptation of uncertainty-aware volume rendering \cite{sakhaee2016statistical} for uncertain SRNs in \cref{sec:ua_vis}. Finally, we conduct ablation studies of the regularization strength $\lambda$ on R\textcolor{black}{MDSRN} as well as the ensemble member count for both \textcolor{black}{MDSRN} and R\textcolor{black}{MDSRN} on \cref{sec:abla_lambda} and \cref{sec:abla_ensszie}.
\vspace{-2 pt}

\subsection{Uncertain Neural Network Architecture Evalaution} \label{sec:usrn_eval}

We quantitatively and qualitatively evaluate  \textcolor{black}{our methods, \textcolor{black}{MDSRN} and R\textcolor{black}{MDSRN}, against Deep Ensemble (DE), Monte Carlo Dropout (MCD), Mean Field Variational Inference (MFVI), and Predicting Variance (PV) reviewed in \cref{sec:background_usrn}}. The uncertain architectures are \textcolor{black}{adapted to} three state-of-the-art feature grid SRNs \textcolor{black}{for} five scalar field datasets with different feature patterns and sparsity. We first detail the experiment settings, and then introduce the evaluation methods and results for the data reconstruction and variance quality tests.

\textbf{Evaluation datasets and base SRNs.} The evaluated datasets include Plume, Nyx, Supernova, Asteroid, and Isotropic. The dimensions of the dataset can be found \textcolor{black}{in the first column }of \cref{tab:acc_uq} containing quantitative results. We apply uncertain network architectures to state-of-the-art feature grid SRNs for SciVis, including fast Volumetric SRN (fV-SRN) \cite{weiss2022fast}, Neural Graphics Primitives (NGP) \cite{wu2023interactive, muller2022instant}, and Adaptively Placed Multi-Grid SRN (APMGSRN) \cite{wurster2023adaptively}. We pair each dataset with the base SRN whose architecture fits the data feature well, as presented in \cref{tab:acc_uq}. For Nyx and Isotropic with dense features over the domain, we use fV-SRN. Plume and Supernova contain features that are uniformly distributed in a subregion, and these features can be properly learned with APMGSRN. As for Asteroid with a higher degree of sparsity, NGP is applied.
Please find the exhaustive quantitative results covering uncertain architectures on all base models for every dataset in the supplemental material.

\textcolor{black}{
\textbf{Baseline uncertain neural network architectures.} Compared methods are reviewed in \cref{sec:background_usrn}.
For SRNs with MCD and MFVI, the MLP decoder is adapted to their methods and the feature grid encoder is unchanged. For DE, independent feature grid SRNs are trained. To implement PV, a feature grid SRN needs an additional output channel for the decoder to predict the variance, and the NLL loss is applied.
}

\textbf{SRN training and hyperparameters.} For each pair of datasets and base SRNs, the uncertain SRNs have the same size from 1\% to 6\% of the data size for a sufficient reconstruction quality.
For all uncertain models except for DE, the same encoder configurations are used, whilst the decoder setup is different to achieve an equivalent size since increasing the grid resolution and feature size can easily result in unbalanced parameter counts.
MCD and PV each has a 3-layer MLP of 128 neurons, and MFVI has 2 layers of 108 neurons.
Both \textcolor{black}{MDSRN} and R\textcolor{black}{MDSRN} have 5 decoders with 2 layers and 64 neurons.
While the 5 members of DE all have a 2-layer decoder with 64 neurons, the encoder capacity is less than other methods for equivalent network sizes.

We implement all models and perform training with PyTorch \cite{paszke2019pytorch} with an NVIDIA A100 GPU and an AMD EPYC 7643 processor.
Training settings are standardized for most models and datasets.
We train most models with the Adam optimizer \cite{DBLP:journals/corr/KingmaB14} and a learning rate of 5.0e-3 which delivers smooth convergences.
For PV, we apply 5.0e-4 as the learning rate following ActiveNeRF \cite{pan2022activenerf} as an existing study of PV SRN, and MFVI also uses 5.0e-4 to avoid unstable training.
We also apply cosine annealing \cite{DBLP:conf/iclr/LoshchilovH17} for learning rate decay with the minimum rate set to 1.0e-7. 
We randomly select one batch of $2^{17}$ or 131,072 coordinate-value pairs in every training iteration for 50000 steps.

For uncertain architectures with hyperparameters, 3 trials of training with different settings are performed.
For MCD, we test the dropout probability $p\in\{0.1,0.15,0.3\}$, and for MFVI, the initial value for the learnable parameter variances has the most influence, and we test values $\in$ \{0.01815, 0.006715, 0.002476\}.
During inference time, 5 stochastic forward passes are used for MCD and MFVI to obtain prediction mean and variance.
For R\textcolor{black}{MDSRN} to be trained with the exponential scheduler defined in \cref{eq:Lvar_scheduler} for the variance regularization, we test $\lambda_{max} \in$ \{10, 30, 100\} with $\lambda_{min}=0$ and $r=500$.
In the following evaluation, we show results from the best trial among the three runs considering both reconstruction and variance quality. Please refer to the supplemental material for the quantitative results of all runs.

\begin{figure*}[t]
\centering
  \includegraphics[alt={volume renderings of error and variance fields}, width=\textwidth]{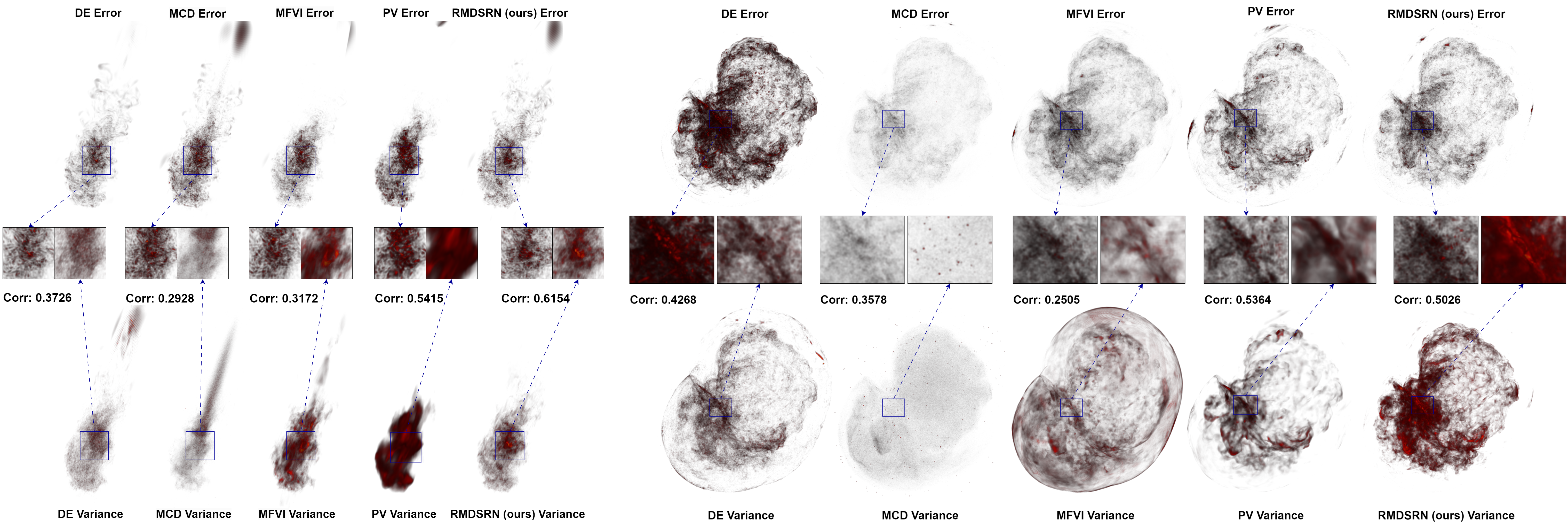}
  \caption{
    Volume renderings of the highest error (top) versus variance (bottom) obtained from uncertain SRNs of Plume on the left and Supernova on the right. R\textcolor{black}{MDSRN} more precisely recovers fine-grained error patterns. While PV attains a high variance-error correlation as a strong competitor, the variance can be oversmooth, potentially overlooking detailed error structures.
  }
  \label{fig:var_eval}
\vspace{-18pt}  
\end{figure*}

\textbf{Data reconstruction evaluation.} To quantify the reconstruction quality of uncertain SRNs, we compute the peak signal-to-noise ratio (PSNR) between the ground truth and the mean reconstructions from uncertain SRNs and present in \cref{tab:acc_uq}.
Results for non-uncertain SRNs are also included with the \quotes{SRN} model label. \textcolor{black}{MDSRN} demonstrates leading PSNR compared to Bayesian methods, DE, and PV in all datasets thanks to the feature-grid-sharing scheme that effectively scales up the member capacity.
\cref{tab:acc_uq} shows \textcolor{black}{MDSRN} can outperform MCD, MFVI, and PV by up to +5 dB in Asteroid with NGP. Furthermore,
the best R\textcolor{black}{MDSRN} among 3 training trials of different $\lambda_{max}$ can further increase the PSNR slightly, making R\textcolor{black}{MDSRN} the most accurate uncertain SRN in the perspective of data-level PSNR.
When comparing the accuracy of R\textcolor{black}{MDSRN} with the SRN without variance quantification, R\textcolor{black}{MDSRN} demonstrates competitive predictive performance close to the non-uncertain base SRN in all datasets except Plume.
This indicates the potential of R\textcolor{black}{MDSRN} in providing quality reconstruction similar to the conventional non-uncertain models while being able to quantify its prediction confidence.
A noteworthy observation for PV is that despite the close performance to R\textcolor{black}{MDSRN} in Nyx, Supernova, and Isotropic, it has the lowest PSNR for Plume and Asteroid.
We observe that PV with NLL training can have a more noisy descent trajectory for the loss compared to other approaches, resulting in a more varying performance between experiments.

For visualization quality comparison, we provide volume rendering images of uncertain-SRN-reconstructed Asteroid and Plume in \cref{fig:Reconstruction}, and each image from SRNs is labeled with image-level metric results comparing against the ground truth including PSNR, structural similarity index measure (SSIM) \cite{wang2004image}, and Learned Perceptual Image Patch Similarity (LPIPS) \cite{zhang2018perceptual}.
We omit \textcolor{black}{MDSRN} in the visual comparison since R\textcolor{black}{MDSRN} can represent the best quality of our methods, and the visual quality difference between their renderings is not significant. R\textcolor{black}{MDSRN} scores the best quantitative results with the lowest perceptual distance plus the highest image PSNR and SSIM, except for Asteroid where PV outperforms R\textcolor{black}{MDSRN} in PSNR.
Nevertheless, with a scrutinization of the enlarged views in Asteroid, R\textcolor{black}{MDSRN} recovers the most detailed feature patterns compared to PV, which is consistent with the SSIM and LPIPS results.
Similarly for Nyx, we can observe that R\textcolor{black}{MDSRN} preserves the shapes and colors of data features highlighted by the transfer function to the greatest extent, and its images look sharper with higher visual fidelity compared to the other renderings that appear fuzzier.

\textbf{Variance quantification evaluation.} To quantitatively evaluate the spatial similarity between the variance and error, we propose to use Pearson correlation (corr) and a modified Jaccard index with spatial tolerance (JI-ST) as the metrics.
Given a variance field and an error field sampled with the same resolution as the data, a higher correlation value suggests high variances and errors are more likely to coincide in the same voxels and vice versa.
Jaccard index\cite{jaccard1912distribution} is computed as the ratio of intersection over union (IoU) between two sets of voxels with one having the highest variances and the other containing the highest errors.
We propose a modified Jaccard index with spatial tolerance (JI-ST) to measure the overlap between the two voxel sets with top variances and errors. When calculating the intersection between the two sets, JI-ST expands the top-error voxel set with one-voxel neighbors of the original members.
This relaxed intersection calculation considers that although the highest variances and errors do not always coincide in the same voxels, if the voxels in the variance set are sufficiently adjacent to those in the error set, the variances are still informative in exposing the weakness of the model in visualization.
For JI-ST, we evaluate the voxel sets with top 1\%- and 5\%-ranked variances and errors.
As the results in \cref{tab:acc_uq} show, PV and R\textcolor{black}{MDSRN} dominate the top one and two variance evaluation metrics.
Although \textcolor{black}{MDSRN} does not demonstrate a clear advantage in the correlation and JI-ST, R\textcolor{black}{MDSRN} improves the similarity of variance to its error considerably and achieves competitive statistics for both metrics compared with PV, surpassing DE and BNNs by a great margin. Since the non-uncertain SRN does not provide variance quantification, the non-applicable metrics are shown as N/A.

Another uncertain SRN evaluation metric is NLL. The mean prediction and variance from the uncertain models are used to parameterize a Gaussian distribution of the prediction, and NLL evaluates how likely the ground truth data are generated from the predicted distributions with smaller values indicating more probable distributions. PV attains the best NLL in most datasets as expected, since NLL is directly minimized during training. After PV, we observe ensemble methods including R\textcolor{black}{MDSRN} and DE attain the highest quality distributions representing the data, despite \textcolor{black}{MDSRN} tending to produce less desirable Gaussians.

To visually evaluate if the variance and error fields correlate well, we compare volume renderings of both fields for Plume (left) and Supernova (right) in \cref{fig:var_eval}. For each dataset, we visualize a percentage of voxels with the highest-ranked variances versus errors to study the effectiveness of variance in revealing regions with the most error-prone predictions.
The data-space correlation is also included between each pair of the error and variance renderings.
For Plume, qualitatively the variance of R\textcolor{black}{MDSRN} captures most precisely the high error regions shown in the cropped view.
Although PV achieves a high variance-error correlation, the variance rendering demonstrates an excessive level of fuzziness that fails to reproduce the detailed error structures for Plume.
As for Supernova, the fuzzy characteristic of the variance from PV is still present, but relatively more spatial structures are revealed than in Plume.
The variance from R\textcolor{black}{MDSRN} displays an overall similar feature structure to the error with more fine-grained patterns captured than other methods.

\begin{figure}[t]% specify a combination of t, b, p, or h for top, bottom, on its own page, or here
  \centering
  \includegraphics[alt={two liner charts showing the error-variance correlation and reconstruction accuracy for compared methods}, width=.95\columnwidth]{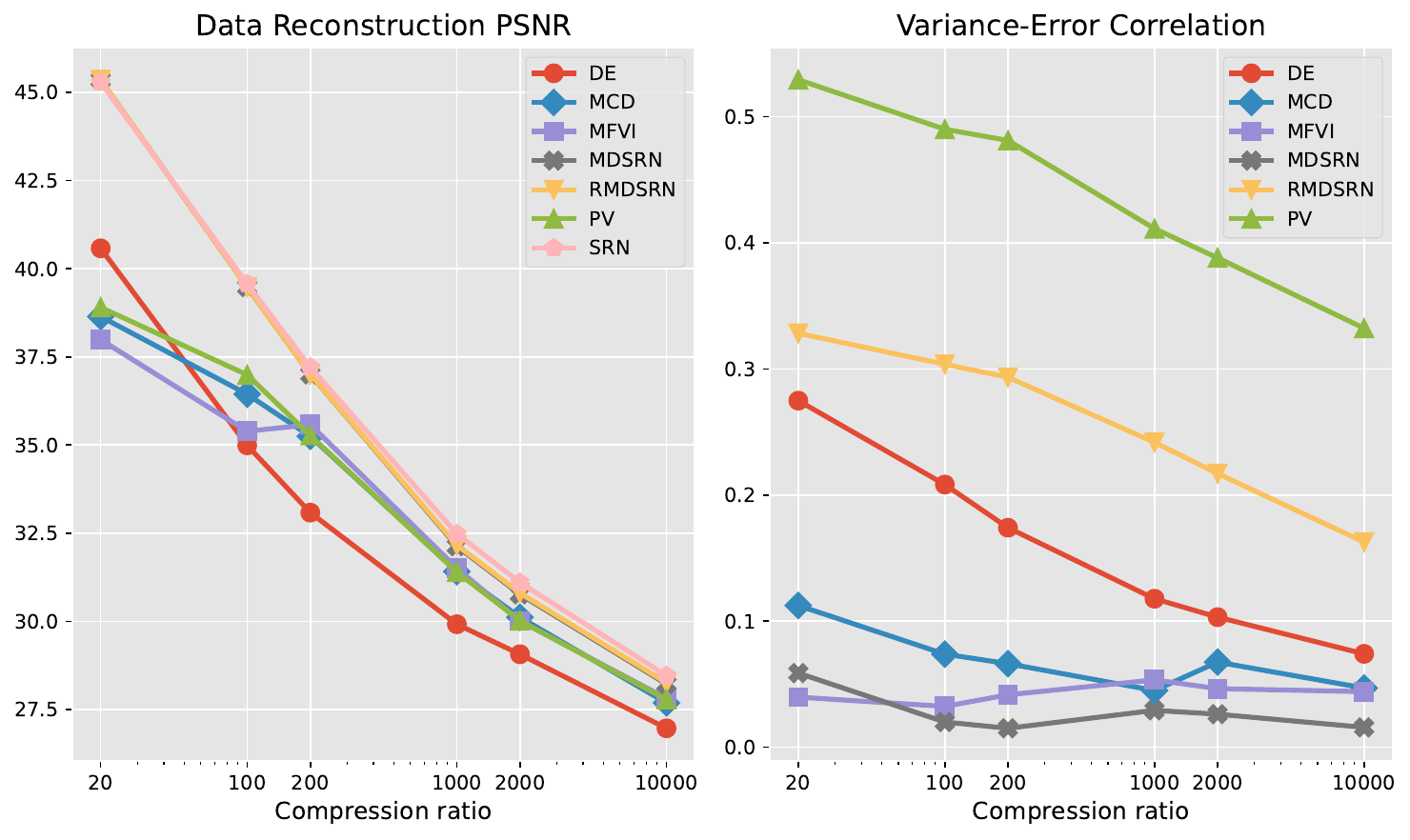}
  \caption{%
  	Data reconstruction and variance-error correlation are evaluated for SRNs across six compression levels. \textcolor{black}{MDSRN} and R\textcolor{black}{MDSRN} lead the reconstruction PSNR in all model sizes, while PV demonstrates more advantageous Pearson correlations between variance and error.
   }
  \label{fig:cross_compress}
  \vspace{-12pt}
\end{figure}

\begin{figure*}[t]
\centering
  \includegraphics[alt={comparison between volume rendering of the mean field versus uncertainty-aware volume rendering with uncertain SRNs}, width=\textwidth]{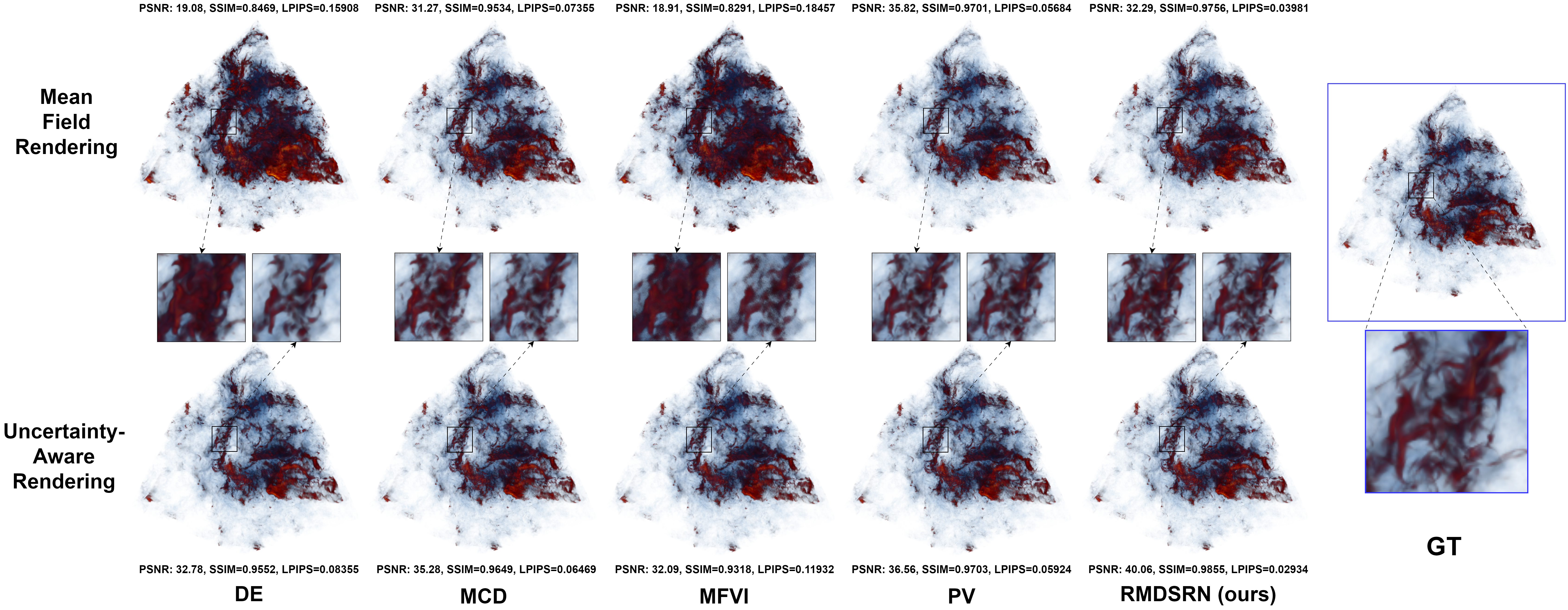}
  \caption{
    Mean prediction volume rendering versus uncertainty-aware volume rendering with our adaptation of statistical DVR \cite{sakhaee2016statistical} for uncertain SRNs. All uncertain SRNs produce more accurate volume renderings with the uncertainty-aware algorithm than rendering the mean, revealing the potential advantage of uncertainty-aware visualization methods for uncertain SRNs.
  }
  \label{fig:ua_rendering}
\vspace{-10pt}  
\end{figure*}

\textbf{Cross-compression-level evaluation.} In addition to the above evaluations comparing uncertain SRNs under a constant size within each dataset and base SRN, we present quantitative results evaluating their performance under six compression ratios from 1/10000 to 1/20 for Isotropic with fV-SRN in \cref{fig:cross_compress}.
Similar to the setup for previous evaluations, we train 3 runs of MCD, MFVI, and R\textcolor{black}{MDSRN} with the mentioned hyperparameter sets and choose the best model with relatively better reconstruction and correlation.
We plot the compression ratio versus reconstruction PSNR and variance-error correlation for applicable models. 
For the per-compression-level PSNR, we can observe R\textcolor{black}{MDSRN} and \textcolor{black}{MDSRN} in gray and yellow achieve leading PSNRs in every compression level, comparable to the SRN without variance estimation, and the accuracy gap for other uncertain SRNs increases with larger model sizes.
Despite a potentially oversmooth variance as shown in the renderings from the variance evaluation, PV consistently outperforms other uncertain models with considerably higher variance-error correlations in the correlation chart. R\textcolor{black}{MDSRN} follows PV with the next best correlation scores. Notably, despite having inferior reconstruction accuracy, DE also exhibits high-quality variance across compression levels over Bayesian methods.

\subsection{Uncertainty-Aware Volume Rendering with Uncertain SRNs} \label{sec:ua_vis}

With uncertain SRNs outputting multiple predictions for any coordinate input, in addition to visualizing the variance as one presented approach for quality-informed visualization, the architectures open the opportunity for the application of probabilistic volume visualization algorithms that work on uncertain data to produce uncertainty-aware visualization \cite{pothkow2010positional, pothkow2011probabilistic, pothkow2013nonparametric, fout2012fuzzy, sakhaee2016statistical, athawale2020direct}.

We present our adaptations and results of the uncertainty-aware volume rendering technique proposed by Sakhaee et al.\cite{sakhaee2016statistical} for uncertain SRNs to directly incorporate the multiple predictions for every sampled location into the final rendered image, beyond rendering only the mean field.
In the statistical direct volume rendering (DVR) framework from Sakhaee et al.\cite{sakhaee2016statistical}, the uncertainty in the data values is integrated into transfer function (TF) classification.
For uncertain data with the value in each spatial location conforming to some probability distribution, they propose to compute the expected color in a sampled location by applying the TF to all possible values and accumulating the colors as well as opacities weighted by the probabilities of the values.
To adapt the statistical DVR framework for uncertain SRNs, we compute the expected color by the sum of colors of each predicted value weighted by the probability of the prediction under the Gaussian distribution parameterized by the mean value and variance. The weights need to be normalized by the total probabilities of all samples to ensure they sum to one before multiplying by the color and opacity.
With this formulation, predictions closer to the mean are given higher weights than others.
This behavior can be justified by the observation that the mean prediction is often more accurate than the samples for both ensemble and BNN methods evaluated, hence samples more similar to the mean are also expected to be higher-quality predictions, and they should contribute more to the expected color and opacity.

We render fV-SRN-based uncertain SRNs for Isotropic with the described adaptation of the uncertainty-aware TF classification method in comparison with the mean prediction renderings in \cref{fig:ua_rendering}.
The models are from the experiments for \cref{sec:usrn_eval}.
We observe the uncertain rendering approach improves the image accuracy considerably for all uncertain SRNs with R\textcolor{black}{MDSRN} showing the most accurate visualization for both mean prediction and uncertainty-aware rendering. By aggregating the post-classification colors instead of classifying the mean prediction with the TF, the features in red displayed in the enlarged views exhibit clearer visibility of the feature structure, and all uncertain SRNs attain higher image accuracy from the quantitative results reported besides the images.

\begin{figure}[t]% specify a combination of t, b, p, or h for top, bottom, on its own page, or here
  \centering
  \includegraphics[alt={line chart showing RMDSRN's performance under different regularization strengths}, width=\columnwidth]{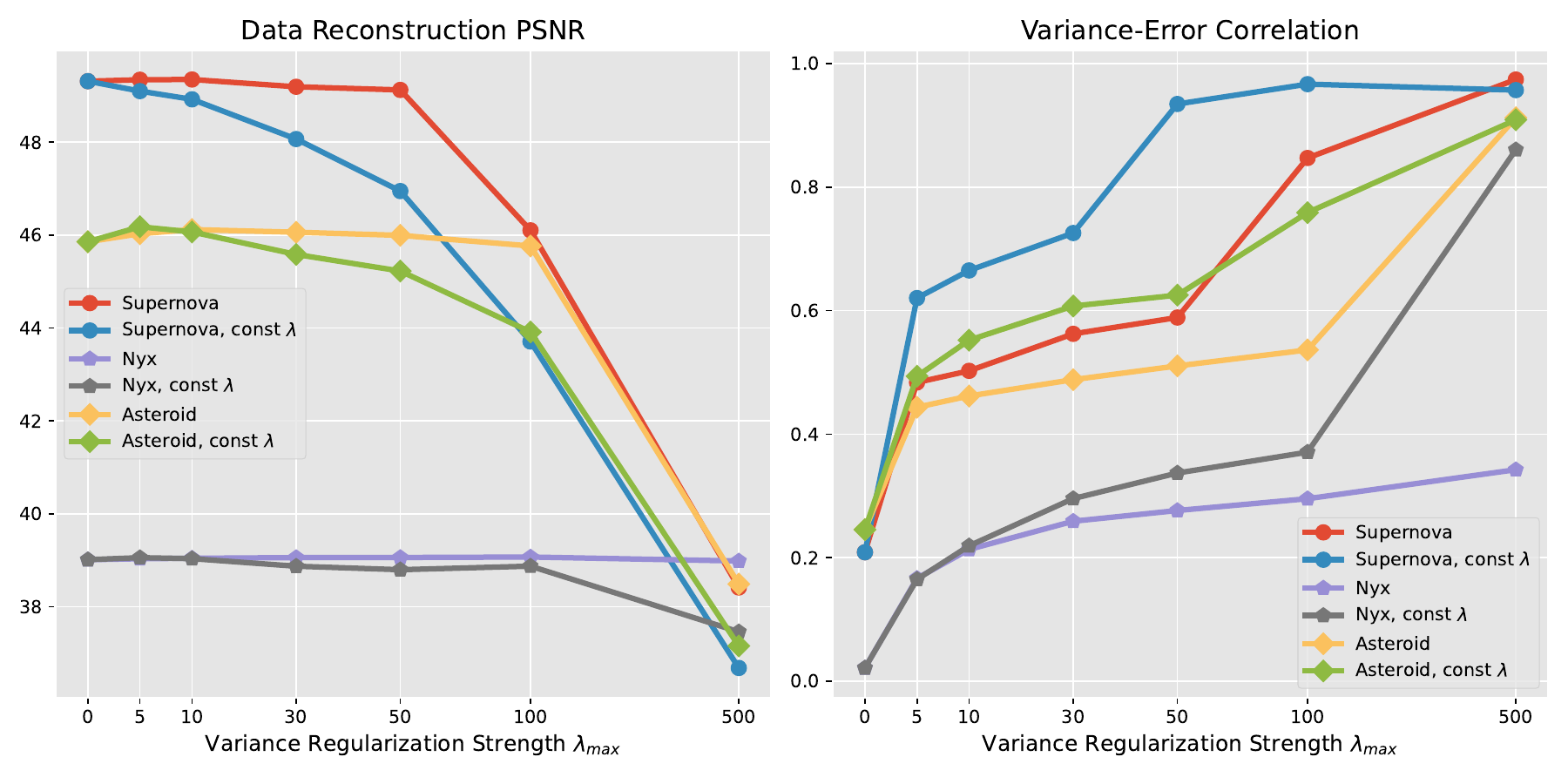}
  \caption{%
  	Effect of varying variance regularization strength $\lambda_{max}$ on R\textcolor{black}{MDSRN} with scheduler tested on Supernova, Nyx, and Asteroid. Results for R\textcolor{black}{MDSRN} with constant $\lambda$ are shown with suffix \quotes{const $\lambda$}. R\textcolor{black}{MDSRN} performs well under wide ranges of $\lambda_{max}$ with the proposed scheduler compared to with a constant $\lambda$.
   }
  \label{fig:lambda_abal}
\vspace{-17pt}
\end{figure}

\subsection{Ablation Study: Regularization Strength and Scheduler} \label{sec:abla_lambda}

The variance regularization endows R\textcolor{black}{MDSRN} with a reliable variance closely correlated to the error down to detailed spatial patterns compared to other approaches, and this performance is tightly coupled with an appropriate regularization strength. We study the behavior of R\textcolor{black}{MDSRN} under different $\lambda_{max}$ in \cref{fig:lambda_abal} with $\lambda_{min}=0$ and $r=500$ as the settings of the $\lambda$ scheduler, compared with an \textcolor{black}{MDSRN} baseline.
% with zero $\lambda_{max}$.
We test on Supernova, Nyx, and Asteroid with network configurations for each dataset the same as in \cref{sec:usrn_eval}. In addition, we test R\textcolor{black}{MDSRN} with constant $\lambda$ throughout training without the proposed scheduler, and the results are labeled with a suffix of \quotes{const $\lambda$}.

Before we compare the performance of R\textcolor{black}{MDSRN} with a constant versus scheduled $\lambda$, we first focus on discussing the results for R\textcolor{black}{MDSRN} with the scheduler plotted in red, yellow, and purple lines in \cref{fig:lambda_abal}. The right subplot shows stronger regularization monotonically increases the variance-error correlation as shown in all datasets, whereas the reconstruction accuracy can degrade after some threshold. Therefore, the ideal $\lambda_{max}$ for the scheduler \textcolor{black}{ranges from zero up to the} maximum value before observing a significant accuracy drop for R\textcolor{black}{MDSRN} to achieve the best quality variance possible while retaining similar reconstruction accuracy as \textcolor{black}{MDSRN}.
For Supernova with \textcolor{black}{MDSRN} scoring a PSNR of 49.31 dB, the best $\lambda_{max}$ for R\textcolor{black}{MDSRN} is between 10 to 30 beyond which the accuracy starts to underperform \textcolor{black}{MDSRN}. This threshold is between 50-100 for Asteroid and beyond 500 for Nyx.
% each with an \textcolor{black}{MDSRN} reconstruction quality of 45.86 dB and 39.01 dB PSNRs respectively.
Comparing the best $\lambda_{max}$ between datasets with different levels of \textcolor{black}{MDSRN} accuracy, we can observe the threshold inversely correlates with model accuracy. In other words, for a given dataset and a base SRN architecture, R\textcolor{black}{MDSRN} with less capacity is expected to be more amenable to a high regularization strength than a larger R\textcolor{black}{MDSRN} that can learn a more accurate reconstruction.

\textcolor{black}{Although the best-performing $\lambda_{max}$ value can differ between training settings, it can be observed that a mild regularization with $\lambda_{max} \in [5, 10]$ works well generally, and there are only slight performance gaps comparing to the optimal $\lambda_{max}$ due to diminishing returns evident in \cref{fig:lambda_abal}, hence we recommend this range as a default setting.}
If it is desired to maximize the benefit of the regularization, an \textcolor{black}{MDSRN} baseline can be trained to guide the $\lambda_{max}$ tuning.
Though the interpretation of PSNR values can vary for different datasets, we recommend trying $\lambda_{max}$ up to 50 if the baseline \textcolor{black}{MDSRN} scores PSNR under 40 dB, or up to 30 for under 45 dB PSNR.
As we have observed the inverse relationship between PSNR and $\lambda_{max}$, a more intelligent scheduler that automatically scales $\lambda$ during training based on the current accuracy might be possible. We leave this to future work and recommend any R\textcolor{black}{MDSRN} to start with the suggested $\lambda_{max}$ between 5 to 10.

We now compare the R\textcolor{black}{MDSRN}s trained with our scheduler versus constant strength, after we have established that a proper regularization should increase the variance quality without degrading the performance of R\textcolor{black}{MDSRN} under the \textcolor{black}{MDSRN} baseline. Taking Asteroid as an example, it is observed in the PSNR chart that R\textcolor{black}{MDSRN} with the scheduler can have $\lambda_{max}$ set between 0 to 100 without hurting the predictive performance. However, with a constant $\lambda$, the results in the green line show the accuracy of R\textcolor{black}{MDSRN} already drops below \textcolor{black}{MDSRN} for strengths greater and equal to 30. The similar trend is manifested in other datasets as well. In particular, a constant $\lambda$ of 5 is already excessive for Supernova. Concluding from the observations, our proposed $\lambda$ scheduling routine is shown to ease the sensitivity of $\lambda$ tuning, such that a wider range of strengths can be applied for better variance quantification over \textcolor{black}{MDSRN} without an adverse side effect on accuracy.
\vspace{-2.3pt}

\subsection{Ablation Study: \textcolor{black}{Decoder Count} } \label{sec:abla_ensszie}

The number of \textcolor{black}{decoder} members can affect the performance of ensemble networks, and we study the performance of our \textcolor{black}{MDSRN} and R\textcolor{black}{MDSRN} with different numbers of member decoders while keeping the total model size constant. We test on Supernova with APGMSRN.

\begin{figure}[t]% specify a combination of t, b, p, or h for top, bottom, on its own page, or here
  \centering
  \includegraphics[alt={line chart comparing RMDSRN's performance with varying ensemble sizes}, width=.9\columnwidth]{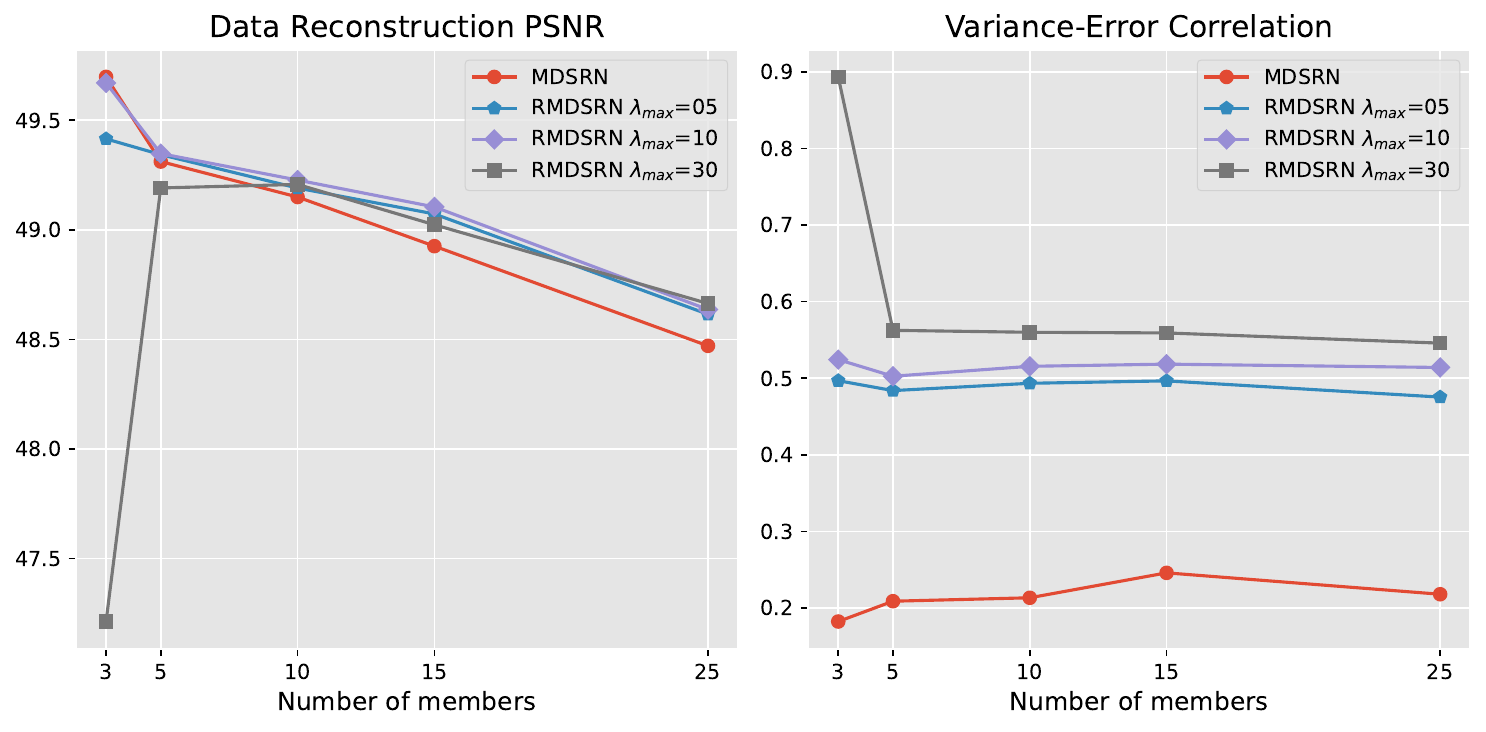}
  \caption{%
  	Effect of varying \textcolor{black}{number of decoders} for \textcolor{black}{MDSRN} and R\textcolor{black}{MDSRN} with different variance regularization strengths.
   }
  \label{fig:enssize}
  \vspace{-15pt}
\end{figure}

As \cref{fig:enssize} reveals, \textcolor{black}{MDSRN} does not exhibit an additional advantage in reconstruction accuracy with more decoders under a constant parameter budget.
This adheres to our empirical observation reported in \cref{sec:esrn} that the most important factor affecting the ensemble accuracy is the capacity of member models.
Nevertheless, the variance quality keeps improving from 3 to 15 decoders.
Following the results, a decoder of 5 for \textcolor{black}{MDSRN} can be a proper choice balancing both accuracy and variance quality. For R\textcolor{black}{MDSRN}, the reconstruction PSNR shows a monotonic decrease for both $\lambda_{max}$ of 5 and 10.
Although R\textcolor{black}{MDSRN} with $\lambda_{max}=30$ shows increased performance from 3 to 5, the lower performance of 3 decoders is likely a result of excessive regularization strength, as the accuracy is substantially lower than \textcolor{black}{MDSRN}.
As for their variance quality, \textcolor{black}{decoder count} is less impactful as the regularization strength and the correlation change in response to different numbers of decoders is minimal. Concluding from the observations, 3 to 5 decoders for R\textcolor{black}{MDSRN} can provide a balanced performance considering accuracy and variance-error correlation.

\section{Limitations and Future Work}

We acknowledge several limitations of our \textcolor{black}{multi-decoder} architecture and regularization with a discussion of possible improvements and future work.
First of all, our \textcolor{black}{MDSRN}, as an ensemble method, requires training multiple member networks for each optimization step. Compared with non-ensemble methods such as BNNs and PV, \textcolor{black}{MDSRN} takes longer to train as shown in \cref{tab:acc_uq}.

Secondly, although \textcolor{black}{MDSRN} achieves great parameter efficiency for better accuracy than DE, PV, and BNNs, the variance quality does not exhibit an advantage as shown in \cref{tab:acc_uq} \textcolor{black}{without the variance regularization}.
\textcolor{black}{The diversity of networks represented by the uncertain model is an important factor affecting the performance \cite{fort2019deep, wilson2020bayesian}, and improving the diversity of the decoders can be future work for better variance quality.}

As studied in \cref{sec:abla_lambda}, regularization strength $\lambda$ has important implications for the accuracy and variance quality of R\textcolor{black}{MDSRN}.
\textcolor{black}{The proposed $\lambda$ scheduler is shown to reduce its sensitivity and provide a default setting that works for most experiments.}
However, $\lambda_{max}$ still needs to be set, and finding the best $\lambda_{max}$ \textcolor{black}{can require} hyperparameter searching. On the other hand, as the proper $\lambda$ appears to inversely correlate with model accuracy, an automatic $\lambda$ scheduler that adjusts the value based on both the training iteration and current prediction accuracy might be able to be developed as future work.

In this work, our utilization of the prediction variance is to facilitate error-aware data reconstruction and visualization,
\textcolor{black}{but it can be exploited for more diverse purposes. We plan to explore the potential of active learning methods with uncertain networks for resource-efficient training of SRNs for large-scale scientific datasets.}

\section{Conclusion}

To equip feature grid SRNs for scientific data with inference time prediction quality assessment, we propose R\textcolor{black}{MDSRN} that provides prediction variance quantification from multiple plausible predictions made to any given input coordinate for error-aware reconstruction of volumetric data. R\textcolor{black}{MDSRN} comprises a parameter-efficient multi-decoder architecture synergized with a novel variance regularization loss for reliable variance estimation well-correlated with prediction error.
Through both quantitative and qualitative evaluations, R\textcolor{black}{MDSRN} demonstrates superior data reconstruction and visualization accuracy as well as competitive variance quality under the same model size across different datasets and compression levels compared to alternative uncertain neural network architectures
shown in \cref{sec:usrn_eval}.
Additionally, we present results of uncertainty-aware volume rendering incorporated to uncertain SRNs, which reveals the potential advantage of leveraging uncertain predictions for a more accurate volume rendering than using the mean field with uncertain SRNs.
As we examine the effect of regularization strength and \textcolor{black}{decoder count} on the performance of \textcolor{black}{MDSRN} and R\textcolor{black}{MDSRN} in the ablation studies, we find our models can achieve satisfactory reconstruction accuracy and variance quality with a relatively small number of members under 5, and our R\textcolor{black}{MDSRN} performs well under a wide range of regularization strengths thanks to the exponential growth scheduler derived for the regularization.

Our work explores an error-aware SRN method realized with uncertain neural network architecture, and we see the potential of continued research on more diverse error-aware SRN approaches and applications to improve the trustworthiness and performance of SRN for scientific visualization in future work.

\onecolumn
\newpage
\twocolumn

\acknowledgments{%
This work is supported in part by the US Department of Energy SciDAC program DE-SC0021360 and DE-SC0023193, National Science Foundation Division of Information and Intelligent Systems IIS-1955764, and Los Alamos National Laboratory Contract C3435.
}

\bibliographystyle{abbrv-doi-hyperref}

\bibliography{template}

\appendix % You can use the `hideappendix` class option to skip everything after \appendix

% \section{About Appendices}
% Refer to \cref{sec:appendices_inst} for instructions regarding appendices.

% \section{Troubleshooting}
% \label{appendix:troubleshooting}

% \subsection{ifpdf error}

% If you receive compilation errors along the lines of \texttt{Package ifpdf Error: Name clash, \textbackslash ifpdf is already defined} then please add a new line \verb|\let\ifpdf\relax| right after the \verb|\documentclass[journal]{vgtc}| call.
% Note that your error is due to packages you use that define \verb|\ifpdf| which is obsolete (the result is that \verb|\ifpdf| is defined twice); these packages should be changed to use \verb|ifpdf| package instead.

% \subsection{\texttt{pdfendlink} error}

% Occasionally (for some \LaTeX\ distributions) this hyper-linked bib\TeX\ style may lead to \textbf{compilation errors} (\texttt{pdfendlink ended up in different nesting level ...}) if a reference entry is broken across two pages (due to a bug in \verb|hyperref|).
% In this case, make sure you have the latest version of the \verb|hyperref| package (i.e.\ update your \LaTeX\ installation/packages) or, alternatively, revert back to \verb|\bibliographystyle{abbrv-doi}| (at the expense of removing hyperlinks from the bibliography) and try \verb|\bibliographystyle{abbrv-doi-hyperref}| again after some more editing.

\end{document}